\documentclass[10pt, conference, compsocconf]{IEEEtran}
\IEEEoverridecommandlockouts

\newtheorem{thm}{Theorem}[section]

\newtheorem{asmp}[thm]{Assumption}

\newtheorem{problem}{Problem}
\newtheorem{exmp}{Example}

\newcounter{mycounter}

\newcommand{\x}{\mathbf{x}}

\makeatletter
\renewcommand*{\@opargbegintheorem}[3]{\trivlist
  \item[\hskip \labelsep{\it\quad  #1\ #2:}] {\it(#3)}\ }
\makeatother

\usepackage{stmaryrd}
\usepackage{graphicx} 
\usepackage{bbm}
\makeatletter
\let\NAT@parse\undefined
\makeatother
\usepackage{hyperref}
\usepackage{epsfig} 
\usepackage{amsmath}
\usepackage{amssymb}
\usepackage{epstopdf}
\usepackage{cite}
\usepackage[noend,ruled,linesnumbered,resetcount]{algorithm2e}
\SetKwComment{Comment}{$\triangleright$\ } {}
\usepackage{multirow}
\usepackage{rotating}
\usepackage{subfigure}
\usepackage{color}
\usepackage{mysymbol}
\usepackage{url}
\usepackage{tabularx,ragged2e,booktabs}
\usepackage[normalem]{ulem}
\usepackage[utf8x]{inputenc}
\usepackage{multirow}

\newcommand{\RNum}[1]{\uppercase\expandafter{\romannumeral #1\relax}}

\usepackage[hang,flushmargin]{footmisc}

  \makeatletter
  \setbox0\hbox{$\xdef\scriptratio{\strip@pt\dimexpr
    \numexpr(\sf@size*65536)/\f@size sp}$}
\newcommand{\scriptveryshortarrow}[1][3pt]{{%
    \hbox{\rule[\scriptratio\dimexpr\fontdimen22\textfont2-.2pt\relax]
               {\scriptratio\dimexpr#1\relax}{\scriptratio\dimexpr.4pt\relax}}%
   \mkern-4mu\hbox{\let\f@size\sf@size\usefont{U}{lasy}{m}{n}\symbol{41}}}}

\makeatother

\begin{document}
\title{Socially-Aware Robot Planning via Bandit Human Feedback
  \thanks{$^\dagger$ These Authors contributed equally to this work. \newline
    This work is supported in part by AFOSR under award \#FA9550-19-1-0169.
}
}

\author{\IEEEauthorblockN{Xusheng Luo$^\dagger$, Yan Zhang$^\dagger$, Michael M. Zavlanos}
\IEEEauthorblockA{\textit{Department of Mechanical Engineering and Materials Science, Duke University, USA}  \\
$\{$xusheng.luo, yan.zhang2, michael.zavlanos$\}$@duke.edu}
}


\maketitle
\begin{abstract}
In this paper, we consider the problem of designing collision-free, dynamically feasible, and socially-aware trajectories for robots operating in environments populated by humans. We define trajectories to be social-aware if they do not interfere with humans in any way that causes discomfort. In this paper, discomfort is defined broadly and, depending on specific individuals, it can result from the robot being too close to a human or from interfering with human sight or
tasks. Moreover, we assume that human feedback is a bandit feedback indicating a complaint or no complaint on the part of the robot trajectory that interferes with the humans, and it does not reveal any contextual information about  the locations of the humans or the reason for a complaint. Finally, we assume that humans can move in the obstacle-free space and, as a result, human utility can change. We formulate this planning problem as an online optimization problem that minimizes the social value of the time-varying robot trajectory, defined by the total number of incurred human complaints. As the human utility is unknown, we employ zeroth order, or derivative-free, optimization methods  to solve this problem, which we combine with off-the-shelf motion planners to satisfy the dynamic feasibility and collision-free specifications  of the resulting trajectories. To the best of our knowledge, this is a new framework for socially-aware robot planning that is not restricted to avoiding collisions with humans but, instead, focuses on increasing the social value of the robot trajectories using only bandit human feedback.

\end{abstract}
\section{Introduction}
Recent advances in onboard computing software and hardware make it appear that the time when robots will populate and operate in the same environments with humans is not too far away. As a result, there has recently been a strong interest in developing planning methods for robots that operate in close proximity to humans~\cite{kruse2013human}. To this date, most of these methods focus on safe planning in dense human crowds, where humans move frequently, and they often rely on standard collision avoidance techniques developed in the robotics literature for this purpose. While safety is indeed a critical specification for any human-robot system, the deployment of robots in environments populated by humans also requires planning methods that can incorporate different human utilities that are not always straightforward to express mathematically. In fact, it is often the case that humans will only provide high-level human feedback on whether they like an event or not, e.g., a robot behavior, without providing relevant contextual information, e.g., their locations in the case of safety. Developing planning methods that can incorporate such unknown human utilities and high-level feedback is necessary to have robots that can efficiently coexist with humans.

In this paper, we consider the problem of planning socially-aware trajectories for robots operating in environments populated by humans. The task assigned to the robot is a basic reach-avoid task. We define a robot trajectory to be social-aware if it does not interfere with humans in any way that causes discomfort. In this paper, discomfort is defined broadly and, depending on specific individuals, it can result from the robot being too close to a human or from interfering with human sight or tasks. Moreover, we assume that human feedback is a bandit feedback indicating a complaint or no complaint on the part of the robot trajectory that interferes with the humans, and it does not reveal any contextual information on the locations of the humans or the reason for a complaint. Finally, we assume that humans can move in the obstacle-free space. As a result, human utilities can change and different robot paths need to be computed for different human configurations. We formulate this planning problem as an online optimization problem that minimizes the social value of the time-varying robot trajectory, defined by the total number of incurred human complaints. Since the human utility function that specifies human feedback is unknown, we can not apply standard gradient-based optimization methods to solve this online optimization problem. Instead, we propose to use derivative-free methods, specifically the zeroth order method, to estimate the gradient by querying humans for feedback on randomly generated trajectories. To obtain a plausible solution, we restrict the number of requests for feedback that can be sent to humans; humans will not provide feedback hundreds of times for the same task. Furthermore, to ensure that the resulting robot trajectories are collision-free and dynamically feasible, we integrate the zeroth order method with an off-the-shelf motion planner that corrects the zeroth-order trajectory to satisfy these additional constraints.

Next, we discuss related work on socially-aware robot navigation, trajectory optimization, and human preference.
\subsection{Socially-Aware Robot Navigation}
To this date, {\it model-based} and {\it learning-based} methods are the two main approaches to socially-aware robot navigation~\cite{chen2017socially}. Model-based methods typically combine multiagent collision avoidance techniques with predictive models of human motion, such as social force models~\cite{ferrer2013robot} or gaussian-processes-based statistical models~\cite{trautman2015robot}. On the other hand, learning-based methods either learn an interaction model which can predict human motion from a human trajectory dataset by means of hand-crafted features~\cite{kuderer2012feature,kretzschmar2016socially}, gaussian processes~\cite{vemula2017modeling}, or neural networks~\cite{alahi2016social}, or directly enforce multiagent cooperative collision avoidance using deep reinforcement learning~\cite{chen2017socially}. As discussed before, common in these methods is that they all focus on human-robot collision avoidance that can be modeled mathematically. Instead, here we consider socially-aware robot planning for human utilities that are possibly unknown and can not be easily described mathematically. Instead, we only assume that high-level human feedback is available that  does not include contextual information about the reasons for a complaint. To the best of our knowledge, this is the first robot planning framework that can incorporate such types of human utilities and feedback.
\subsection{Trajectory Optimization}
Numerical optimization methods for motion planning can be broadly classified into two categories, namely, {\it derivative-based} and {\it derivative-free}. Derivative-based methods require a smooth and differentiable objective function and typically solve non-linear motion planning problems by leveraging the first- and second-order derivatives. These methods include, but are not limited to, the well-known motion planners CHOMP~\cite{ratliff2009chomp}, TrajOpt~\cite{schulman2013finding}, KOMO~\cite{toussaint2017tutorial}, and RieMo~\cite{ratliff2015understanding}. However, derivative-based methods can not handle non-differentiable utility functions, such as the human utilities considered in this work. Such non-differentiable utilities can instead be handled by derivative-free planning methods. A type of derivative-free planning methods that have been widely used in the robotics literature are  the sampling-based methods, such as kinodynamic RRT$^*$~\cite{webb2013kinodynamic}, which builds graphs  by randomly exploring the configuration space and optimizing an objective along the way. Alternatively, population-based optimization methods can be used for motion planning with non-differentiable objectives, such as the partical swarm optimization method~\cite{kim2015trajectory}, that moves a swarm of particles in the search space in search of better robot trajectories. Another popular derivative-free method is STOMP~\cite{kalakrishnan2011stomp} that improves on the current trajectory by applying  Monte Carlo simulation to sample a series of noisy trajectories and explore the trajectory space. Although derivative-free planning methods, such as those proposed in~\cite{webb2013kinodynamic,kim2015trajectory,kalakrishnan2011stomp}, can handle non-differentiable objectives, they still require these objectives to be known and mathematically expressed. This is not the case in the socially-aware planning problem considered here, where human utilities are unknown and only high-level context-unaware human feedback is available. What makes it possible to utilize such human feedback for robot planning is the proposed zeroth-order optimization framework that is combined with off-the-shelf planners such as the above.
\subsection{Human Preference Trajectory Optimization}\label{sec:humanpreference}
Perhaps most relevant to the method proposed here is recent work on trajectory optimization using human preferences. For example, in~\cite{mcgill2016low} task-specific human preferences are incorporated into  a motion planner, assuming that these human preferences are formulated as simple convex functions. {Similarly, in~\cite{jain2015learning}, human preferences are used to iteratively make small improvements to trajectories returned by robots.} Instead, here we assume that  human feedback is a bandit feedback and the human utility function that governs this feedback is unknown. Following a different approach,~\cite{menon2015towards} adopts a data-driven method to account for human preferences. A set of features regarding the trajectory is designed and a classifier is learned to predict human rating on a new trajectory. Similarly,~\cite{daniel2015active} learns an action policy to tackle grasping tasks by actively querying a human expert for ratings. More broadly, preference-based reinforcement learning~\cite{wirth2017survey} focuses on human preference for pairwise comparison between trajectories. Either a policy that satisfies human preference is learned directly~\cite{wilson2012bayesian,busa2014preference}, or a qualitative preference model capturing a human preference relation or a quantitative utility function estimating a human evaluation criterion are learned and then used to optimize the policy~\cite{furnkranz2012preference,christiano2017deep,sadigh2017active,pinsler2018sample}. While preference-based learning methods can theoretically be used to solve the socially-aware planning problem with bandit human feedback proposed in this paper, their application is not straightforward. First, they can only be applied to stationary problems where the utilities of humans do not change with time. This is because a learned policy for one preference model can not be easily adapted to another preference model corresponding to different human utilities. However, even in the stationary case, the amount of data required by learning-based methods to learn an optimal policy is typically very large. This would make the application of such methods prohibitive in practice,  since humans would need to be queried for feedback  a prohibitively large number of times. Our proposed framework does not suffer such limitations. Extensive numerical simulations show that it can easily handle changing human utilities and the number of queries for human feedback is low.

The rest of the paper is organized as follows. In Section~\ref{sec:preliminary}, we provide a brief overview of the zeroth order optimization method and in Section~\ref{sec:problemformulation} we present the problem formulation. In Section~\ref{sec:solution} we describe our proposed algorithm for the social aware robot planning problem. Numerical experiments are presented in Section~\ref{sec:simulation}.

\section{Preliminaries}\label{sec:preliminary}
In this section, we provide a short background on zeroth-order optimization since it constitutes the foundation  upon which our proposed method is developed. The zeroth order method~\cite{nesterov2017random,gasnikov2017stochastic} is a derivative-free optimization framework  that optimizes an objective function  based only on the function evaluations, which is a preferred  method when the objective function is unknown or noisy, or the gradient is difficult to calculate. Consider an unknown objective function $g:\ccalX \to \mathbb{R}$ defined on a $d$-dimensional domain $\ccalX \subset \mathbb{R}^d$. To minimize the objective $g$, the zeroth order method estimates the gradient at the current iterate $x_k$ by randomly perturbing $x_k$ and then querying for function evaluations one or multiple times. For example, the following gradient estimate queries for function evaluations twice,
\begin{align}\label{equ:gradientestimate}
  \nabla g_x (x_k ) \approx \frac{ n_x ( g (x_k + \delta \mathbf{u}) - g(x_k - \delta \mathbf{u}))} {2\delta} \mathbf{u},
\end{align}
where $n_x$ is the size of the decision variable $x$, $\mathbf{u}$ is a random vector sampled from a uniform distribution in a unit sphere or a Gaussian distribution in $\mathbb{R}^{n_x}$ and $\delta>0$ is the exploration parameter. As discussed in \cite{nesterov2017random,gasnikov2017stochastic}, the expression  in  \eqref{equ:gradientestimate} is an unbiased stochastic gradient estimate of the smoothed function $\mathbb{E}_u [ g(x_k + \delta \mathbf{u}) ]$. Therefore, when the zeroth order method updates the current iterate $x_{k}$ using \eqref{equ:gradientestimate}, it in fact minimizes the smoothed objective function $\mathbb{E}_\mathbf{u} [ g(x_k + \delta \mathbf{u}) ]$, whose distance to $g(x_k)$ is bounded by an amount proportional to the exploration parameter $\delta$, see \cite{nesterov2017random,gasnikov2017stochastic}. After obtaining the estimated gradient, the next iterate is computed in a usual way as
\begin{align}
x_{k+1}  = x_k - \eta \nabla g_x(x_k),
\end{align}
where $\eta > 0$ is the  step size.
\section{Problem Definition}\label{sec:problemformulation}
Consider an environment $\ccalW \subset \mathbb{R}^w, w\in\{2,3\} $, let $\ccalO=\{o_1,\ldots,o_m\}$ denote the set of $m$ obstacles in $\ccalW$, and assume that the obstacle-free space $\ccalW\setminus \ccalO$ is populated by humans that are distributed possibly in a spatially uneven way. Assume also that {the model of human dynamics is unknown} and human movement is random,  making it unlikely to predict the moving pattern. Given a group of $p$ humans in the environment, let $\ccalL_{T_i} = \{l_{1,T_i}, \ldots, l_{p,T_i}\}$ be the set that collects their locations at discrete time instants $T_i, i \in \{0, \ldots, K\}$. We make the following assumption on the motion  pattern of the humans.
\begin{asmp}\label{asmp:move}
There exists a constant $\delta T>0$ such that $T_{i+1}-T_i >\delta T$ for all $i=0,\ldots,K-1$. Moreover, during every time instant $T_{i+1}$, humans $k\in\{1,\ldots,p\}$ move instantaneously between consecutive locations $l_{k, T_i}$ and $l_{k, T_{i+1}}$, and they remain stationary during the time interval $(T_i,T_{i+1})$.
\end{asmp}

The existence of the bound $\delta T>0$ in Assumption~\ref{asmp:move} ensures that the times when humans move in the environment are not arbitrarily close to each other. Since humans also move instantaneously, Assumption~\ref{asmp:move} captures situations where human motion is infrequent and sporadic.  For example, it models typical work environments, where humans remain around certain locations for a while and then quickly move to other locations to perform subsequent tasks.

Consider also a robot navigating in the obstacle-free environment $\ccalW\setminus \ccalO$ from a starting location $x_0$ to a target location $x_g$. Let $q\in \mathcal{Q}$ denote the state of the robot that includes the robot's location, orientation, etc, where $\mathcal{Q}$ is the feasible state space. Let  also $u\in \mathcal{U}$ denote the control input of the robot, where $\mathcal{U}$ is the set of admissible control inputs. Moreover, let $t_j \in [T_i, T_{i+1})$ for $j\in\{0,1,\ldots,n\}$, denote a sequence of  time instants defined over the time intervals that humans remain stationary, and define the discrete-time dynamics of the robots during this time interval as
\begin{align}\label{equ:generalmodel}
  q_{t_{j+1}}=f(q_{t_j},u_{t_j}).
\end{align}
Finally, define a discrete robot trajectory during the time interval $[T_i,T_{i+1})$ by the sequence of robot positions $\x_{T_i} = x_{t_0}, x_{t_1}, \ldots,  x_{t_n}$, where $x_{t_j}\in \ccalW\setminus \ccalO$ is the position component of $q_{t_j}$ and $t_j \in [T_i, T_{i+1})$ for all $j \in \{0, \ldots, n\}$. Particularly, $x_{t_0}=x_0$ is the starting location and $x_{t_n}=x_g$ is the target location.  In what follows, with slight abuse of notation, we write $\x_{T_i} = [x^\intercal_{t_0}, x^\intercal_{t_1}, \ldots, x^\intercal_{t_n}]^\intercal$.

Our goal is to design a collision-free, dynamically feasible, and socially-aware trajectory for the robot that does not interfere with humans in any way that causes discomfort.  In this paper, discomfort is defined broadly and, depending on specific individuals, it can result from the robot being too close to a human, or from interfering with human sight or tasks. To do so, in what follows, we propose a novel way to incorporate human feedback in the robot planning process. Specifically, we make the following assumptions on the feedback provided by humans.
\begin{asmp}\label{asmp:humanfeedback}
Human feedback is a bandit feedback indicating a complaint or no complaint on the that part of the robot trajectory that falls in human vicinity. Moreover, human feedback does not reveal any contextual information about  the locations of the humans or the reason for a complaint. Finally, the human utility function does not change during time intervals $(T_i,T_{i+1})$, but it can change at time instants $T_i$ when humans move.
\end{asmp}

We model human feedback using an implicit  human-and-time-dependent utility function $h_{k,T}: \x_T \mapsto \{0,1\}$ which indicates the preference of individual $k$ regarding the trajectory $\x_T$ at time instant $T$, where we drop the subscript $i$ from  $T_i$ when the notation is clear. Particularly, the function $h_{k,T}$ returns 1 if individual  $k$ complains about the trajectory at time instant $T$, otherwise it returns 0. Dependence of $h_{k,T}$ on time $T$ captures the fact that an individual's feedback on a trajectory may vary over time. For instance, humans may become more or less tolerant to a robot moving in their vicinity, depending on the type of their ongoing work. Essentially,  we treat each individual as a black box and respect their preferential heterogeneity by means of distinct functions $h_{k,T}$. We capture the {\it social value} of  the trajectory $\x_T$ by the total number of complaints incurred by that trajectory, defined as $h_T(\x_T) = \sum_{k=0}^p h_{k,T}(\x_T)$. Then, a trajectory $\x_T$ is said to be {\it socially-aware} if it incurs zero complaints, i.e, $h_T(\x_T) = 0$. {In what follows we avoid relying on models of human feedback to achieve socially-aware planning~\cite{mcgill2016low,menon2015towards,sadigh2017active}, since such models can differ across humans and vary over time.}

On the other hand, we define a collision between the trajectory $\x_T$ and an obstacle $o_j$ if a line segment between consecutive waypoints in $\x_T$ intersects the obstacle $o_j$. Then the trajectory $\x_T$ is {\it collision-free} if it never collides with any obstacle.\footnote{Note that, in this paper, collision avoidance is defined in terms of obstacles. Collisions with humans are captured by human complaints.} Let $\ccalC$ be a set that collects all collision-free discrete trajectories in the environment $\ccalW\setminus \ccalO$.
Finally, we define a {\it dynamically-feasible} trajectory to be one that respects the dynamical robot constraints~\eqref{equ:generalmodel}, and collect all dynamically-feasible discrete trajectories  in the environment $\ccalW\setminus \ccalO$ in the set $\ccalD$. Then, the socially-aware robot planning problem considered in this paper can be formulated as
\begin{equation}
\begin{aligned}\label{equ:offline}
  \min_{{\x}_{T_0},  \ldots, {\x}_{T_{K-1}}} &  \quad \sum_{i=0}^{K-1} h_{T_i}({\x}_{T_i}) \\
  \text{s.t.} \quad \quad& \quad \x_{T_i} \in \ccalC \cap \ccalD, \, \forall i \in \{0, \ldots, K-1\}.
\end{aligned}
\end{equation}
Problem~\eqref{equ:offline} is an {\it offline} optimization problem and it can be solved after time instant $T_{K-1}$ assuming that all utility functions $h_{T_i}$ are known in hindsight. However, if the human utilities are unknown and a new socially-aware trajectory has to be designed during every time interval $[T_{i}, T_{i+1})$, problem~\eqref{equ:offline} needs to be solved in an {\it online} fashion. We can do this by solving the following optimization problem during every time interval $[T_i, T_{i+1})$,
\begin{align}\label{equ:minfeedback}
  \min_{{\x}_{T_i}} &  \quad h_{T_i}({\x}_{T_i})\quad  \text{s.t.} \quad \x_{T_i} \in \ccalC \cap \ccalD
\end{align}
 Note that the objectives in problem~\eqref{equ:minfeedback} are {time-varying} due to the varying human locations and feedback. Based on problem~\eqref{equ:minfeedback}, the socially-aware robot planning problem considered in this paper can be formulated as follows.

\begin{problem}\label{prob:1}
  Let Assumptions~\ref{asmp:move} and~\ref{asmp:humanfeedback} hold. Given a pair of starting and target locations and  a sequence of discrete time instants $T_0, T_1, \ldots, T_K$, with $ K\in \mathbb{N}$, design a sequence of collision-free, dynamically-feasible, and socially-aware trajectories ${\x}_{T_0}, {\x}_{T_1}, \ldots, {\x}_{T_{K-1}}$ each of which solves the optimization problem~\eqref{equ:minfeedback}.
\end{problem}

The solution of an online optimization problem  should track the solution of its offline counterpart~\cite{shahrampour2017distributed}.
As such, we can measure the  quality of the  sequence of trajectories returned by the solution of  Problem~\ref{prob:1} using the {\it cumulative regret} over the whole time period defined as
\begin{align}\label{equ:regret}
  R_K  = \sum_{i=0}^{K-1} h_{T_i}(\tilde{\x}_{T_i}) - \sum_{i=0}^{K-1} h_{T_i}(\x^*_{T_i}),
\end{align}
where $\tilde{\x}_{T_0}, \ldots, \tilde{\x}_{T_{K-1}}$ is the computed solution sequence of problem~\eqref{equ:minfeedback} and $\x^*_{T_0}, \ldots, {\x}^*_{T_{K-1}}$ is the minimizer sequence of problem~\eqref{equ:offline} that incurs the least number of complaints if the functions $h_{T_i}$ are known in hindsight. The cumulative regret is a dynamic regret since the optimal trajectory $\x^*_{T_i}$ varies due to human motion and the variations in the human feedback function $h_{T_i}$ over time. As shown in \cite{mokhtari2016online}, the best known bound for the dynamic regret \eqref{equ:regret} grows at the same rate at which  the optimal trajectory $\x^*_{T_i}$ varies over time.  Therefore, the cumulative utility \eqref{equ:regret} shall at best grow at the same rate at which the human positions and preferences change. Note that since the human utility $h_{T_i}(\x_{T_i})$ is unknown, we can not employ standard gradient-based optimization methods to solve Problem~\ref{prob:1}. Instead, we employ the zeroth order optimization, discussed in Section~\ref{sec:preliminary}, to solve this online problem.

\section{Socially-Aware Robot Planning}\label{sec:solution}
In this section, we first reformulate  Problem~\ref{prob:1} by manipulating the constraint set in~\eqref{equ:minfeedback} and then develop a novel method that combines  zeroth order  optimization with Model Predictive Control (MPC)~\cite{rawlings2000tutorial,grune2017nonlinear,allgower2012nonlinear} to solve the socially-aware robot planning Problem~\ref{prob:1}. Specifically, let $\mathcal{F}(\x)$ be the set of collision-free and dynamically-feasible trajectories that are in the neighborhood of a reference trajectory $\x$ and define the problem

\begin{align}\label{equ:minfeedbackMset}
    \min_{\x_{T_i}} &\quad h_{T_i}(\x_{T_i}) \quad \text{s.t.} \quad \x_{T_i} \in \ccalF(\x_{T_i}).
\end{align}
Note that problems~\eqref{equ:minfeedback} and~\eqref{equ:minfeedbackMset} are equivalent. To see this, observe that if $\x_{T_i}^*$ is the optimal solution of~\eqref{equ:minfeedback} then it is collision-free and dynamically feasible. Therefore, it belongs to the set $\mathcal{F}(\x_{T_i}^*)$. Since~\eqref{equ:minfeedback} and~\eqref{equ:minfeedbackMset} share the same cost function, this means that $\x_{T_i}^*$ is also the optimal solution of~\eqref{equ:minfeedbackMset}. On the other hand, if $\x_{T_i}^*$ is the optimal solution of~\eqref{equ:minfeedbackMset} then it is collision-free and dynamically feasible. Therefore, it satisfies the constraints of problem~\eqref{equ:minfeedback}, which means that $\x_{T_i}^*$ is also the optimal solution of~\eqref{equ:minfeedback}.

Computing the set $\ccalF(\x)$ is generally difficult. Instead, we can compute a single trajectory in $\mathcal{F}(\x)$ using off-the-shelf motion planners, such as MPC.{
Let $m(\x)$ denote such a trajectory, which satisfies $m(\x)\in \mathcal{F}(\x)$. Then, a sufficient condition for feasibility of problem~\eqref{equ:minfeedbackMset} is that $\x_{T_i}=m(\x_{T_i})$. To enforce this constraint, we consider $m(\x_{T_i})$ as a target trajectory that $\x_{T_i}$ needs to track. Moreover, we query humans for feedback on the trajectory $m(\x_{T_i})$ rather than the trajectory $\x_{T_i}$. This is because $\x_{T_i}$ may not be collision-free or dynamically feasible, while $m(\x_{T_i})$ is. Therefore, it is not meaningful to request human feedback on an infeasible trajectory. With the above modifications, we obtain the following {penalty-based} reformulation of problem~\eqref{equ:minfeedbackMset} as
\begin{align}\label{equ:minfeedbacpenaltyapprox}
  \min_{\x_{T_i}} &\quad h_{T_i}(m(\x_{T_i})) + \rho\, \| \x_{T_i} - m(\x_{T_i})\|,
\end{align}
where $\rho$ is a positive penalty parameter and $\|\x_{T_i}-m(\x_{T_i})\|$ measures the tracking error between $\x_{T_i}$ and the motion planner. Note that~\eqref{equ:minfeedbacpenaltyapprox} is not necessarily equivalent to~\eqref{equ:minfeedbackMset}. This is because there is no guarantee that a motion planner can return a trajectory $m(\x_{T_i}^*)$ that satisfies $\x_{T_i}^*=m(\x_{T_i}^*) \in \ccalF(\x^*_{T_i})$, where $\x_{T_i}^*$ is the optimal solution of~\eqref{equ:minfeedbackMset}. Note also that the solution of~\eqref{equ:minfeedbacpenaltyapprox} returns a trajectory $\x_{T_i}$ but the robot executes the trajectory $m(\x_{T_i})$ that is guaranteed to be collision-free and dynamically feasible. {Finally, observe that in~\eqref{equ:minfeedbacpenaltyapprox}, the goal is to minimize human complaints, but other objectives such as control effort can also be incorporated in this formulation. For example, the MPC trajectory $m(\x_{T_i})$ in~\eqref{equ:minfeedbacpenaltyapprox} can be easily constructed in a way that minimizes such additional objectives; see, e.g., Section~\ref{sec:mpc}.}}.



\begin{figure}[t]
  \centering
  \includegraphics[width=\linewidth]{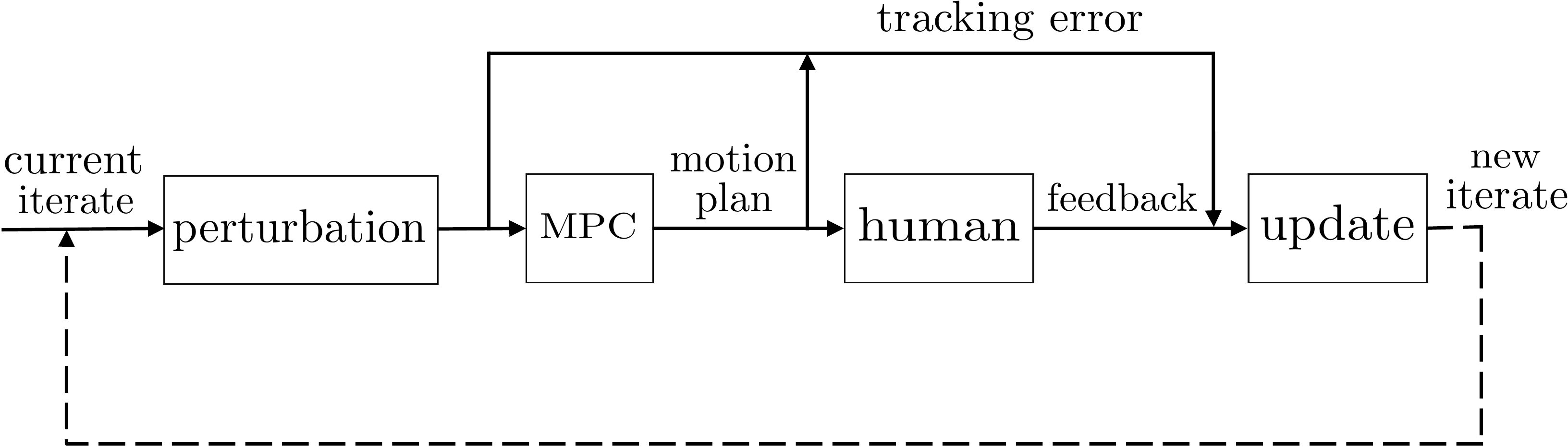}
  \caption{Zeroth order method for socially-aware robot planning}\label{fig:framework}
\end{figure}

As discussed in Section~\ref{sec:problemformulation}, since the human utility function $h_T$ is unknown, we need to utilize  the zeroth order method to solve problem~\eqref{equ:minfeedbacpenaltyapprox}, which we combine with an MPC motion planner to obtain the trajectories $m(\x_{T_i})$ in~\eqref{equ:minfeedbacpenaltyapprox} as follows. At every iteration of the proposed zeroth order method, the current trajectory iterate is perturbed and then rectified by the MPC motion planner so that it is collision-free and dynamically feasible. Then, the humans are queried for feedback on the perturbed trajectories and this feedback along with the tracking error is used by the zeroth order method to estimate the gradient and update the current trajectory iterate; see Section~\ref{sec:preliminary}. The proposed algorithm is illustrated in Figure~\ref{fig:framework}. In what follows, we discuss the proposed zeroth order method, the MPC motion planning problem, and the integrated system in detail.

\subsection{Zeroth Order Optimization Method}\label{sec:zeroth}
The advantage of the proposed zeroth-order method is that it can be used to incorporate bandit human feedback in motion planning. Since only the value of the human utility function $h_{T}(m(\x_{T}))$ is accessible, the proposed zeroth order method uses these values to estimate the gradient of human feedback as
\begin{equation}
        \label{eqn:ZerothOrderMethod}
     { \nabla h_{T}(m(\x^k_{T})) }\approx \frac{n_{\x} \left( h_T ( m(\x_T^{k,+}) ) - h_T (m(\x_T^{k,-}))  \right)} {2\delta} \mathbf{u},
\end{equation}
where $\x_T^{k,\pm} = \x_T^k\pm \delta \mathbf{u}$, $\x^k_{T}$ is the $k$-iterate of $\x_T$, $m(\x_{T})$ is the trajectory returned by is the MPC motion planner that will be introduced in the next section, $n_\x$ is the size of vector $\x_T$, and $\mathbf{u}$ is a random vector of unit length. Given the vector $\mathbf{u}$, the trajectories $m(\x^k_T + \delta \mathbf{u})$ and $m(\x^k_T - \delta \mathbf{u})$ are deterministic.

A main limitation of zeroth-order methods in general is that their  convergence is slow, especially when the problem dimension is high~\cite{nesterov2017random,gasnikov2017stochastic}. This is the case here as the number of waypoints $n_\x$ in the trajectory $\x_T$ can be very large. To address this limitation, we propose local updates of the trajectory $\x_T$, restricted to those waypoints that are directly affected by human complaints. This is a reasonable assumption, since humans are more likely to complain on that part of the trajectory that causes them discomfort rather than on the whole robot trajectory, which may not even be  fully observable. To develop the proposed local update scheme, we replace the utility functions $h_{k,T}(m(\x_T))$ by functions $h_{k,T}([m(\x_T)]_{\mathcal{S}^{k,T}})$, where the index set $\mathcal{S}^{k,T}$ captures the local awareness of human $k$, that is, those waypoints in the whole trajectory $m(\x_T)$ that directly affect human $k$ at time instant $T$.  In this case, to minimize the utility function $h_T(m(\x_T)) = \sum_{k=0}^p h_{k,T}([m(\x_T)]_{\mathcal{S}^{k,T}})$, it is sufficient to only update the waypoints in the set $\mathcal{S}_p \triangleq \cup_{k} \mathcal{S}^{k,T}$ according to the modified zeroth order update
\begin{equation}\label{eqn:ModifiedZerothOrder}
      \nabla h_{T}^{\mathcal{S}_p}(m(\x^k_{T})) \approx \frac{ w|\mathcal{S}_p|\left( h_T ( m (\x_T^{k,+}) ) - h_T (m (\x_T^{k,-}))  \right)} {2\delta} \mathbf{u}^{\mathcal{S}_p},
\end{equation}
where $\x_T^{k,\pm} = \x_T^k\pm \delta \mathbf{u}^{\mathcal{S}_p}$,  $w\in\{2,3\}$ is the dimensionality of a waypoint, $|\mathcal{S}_p|$ is the cardinality of the set $\mathcal{S}_p$ and the vector $\mathbf{u}^{\mathcal{S}_p}$ is computed by setting the entries of $\mathbf{u}$ that are not in the set ${\mathcal{S}_p}$ to be $0$ and then nornalizing, i.e., $\mathbf{u}^{\mathcal{S}_p} = \mathbf{u}^{\mathcal{S}_p}/\|\mathbf{u}^{\mathcal{S}_p}\|$. Since $|\mathcal{S}_p| \ll n_\x$, the convergence speed of the algorithm can be greatly improved. Similarly, the gradient of the tracking error between the zeroth-order iterate and the MPC motion plan can be obtained as
\begin{align}
  \nabla e^{\mathcal{S}_p} (\x_T^k) \approx  \frac{ w|\mathcal{S}_p|\left( e(\x_T^{k,+} ) - e(\x_T^{k,-} ) \right) } {2\delta} \mathbf{u}^{\mathcal{S}_p},
\end{align}
where $e (\x_T^k) = \|\x_T^k - m(\x_T^k) \|$ denotes the tracking error.
Note that $\nabla e^{\mathcal{S}_p}\left(\x_T^k\right)$ can also be computed using finite differences since this objective is known. In what follows, we discuss how to compute the set $\ccalS_p$.

Since each trajectory needs to connect the starting and target locations of the robot, we assume that the first and last waypoints of the trajectory $\x_T$ do not affect any human and, therefore, they are not included in the set ${\mathcal{S}_{p}}$. All other locations in the free space, if they are traversed by the trajectory $\x_T$, can possibly cause discomfort to humans. In what follows, we discuss two forms of human feedback, thus, resulting in two ways to decide the set $\mathcal{S}_p$. {\it (i)} Humans can only send complaints without revealing which waypoints affect them.  In this case, the set $\mathcal{S}_p$ is the full trajectory and Alg.~\ref{alg:zero} perturbs the whole trajectory $\x_T$. We refer to this as the full perturbation scheme. {\it (ii)} Humans send complaints and also report the part of the trajectory that affects them. In this case, the set $\ccalS_p$ contains those waypoints indicated by the human as well as a random number of waypoints before and after the reported sequence of waypoints. This provides more room to the zeroth-order method to compute trajectories that can be easier modified by the motion planner. Moreover, these additional waypoints can account for human errors in reporting the waypoints that cause them discomfort. We refer to this as the local perturbation scheme.
\subsection{Model Predictive Control (MPC)}\label{sec:mpc} MPC has long been used to plan  collision-free robot trajectories that track predefined reference paths~\cite{lapierre2007combined,yoon2009model,gao2011predictive}. We adopt the approach proposed in~\cite{yoon2009model}. Specifically, given a reference trajectory $x_{t_0}, x_{t_1}, \ldots, x_{t_n}$ returned by the zeroth order method that can be dynamically infeasible or can collide with obstacles, our goal is to compute the optimal control sequence $\{u^*_{t_j,k}\}_{k=0}^{N_{m}-1}$, at each time instant $t_j, j\in \{0, \ldots, n-1\}$,  by solving the following optimal control problem over a horizon of length $N_m$:
\begin{subequations}
  \begin{align}
    & \{u^*_{t_j, k}\}_{k=0}^{N_{m}-1}  = \argmin J\left(x_{t_j}, \{u_{t_j, k}\}_{k=0}^{N_{m}-1}\right) \nonumber \\
 &= \mu \sum_{k=0}^{N_m-1} J^{\text{obs}}(x_{t_j, k}, x_{t_j, k+1}) + \frac{1}{2} \tilde{x}_{t_j, N_m}^\intercal P\tilde{x}_{t_j, N_m}  \nonumber \label{equ:obj} \\
   &+ \sum_{k=0}^{N_m-1} \left(\frac{1}{2} \tilde{x}_{t_j, k}^\intercal Q\tilde{x}_{t_j,k} + \frac{1}{2} u_{t_j, k}^\intercal R u_{t_j,k}\right)
  \end{align}
  \begin{align}
    & \text{s.t.} \quad   q_{t_j, k+1,} = f(q_{t_j, k}, u_{t_j, k})\\
  &\quad \quad\, u_{t_j, k} \in \ccalU, \;\forall k \in \{0, 1, \ldots, N_m-1\}
\end{align}
\end{subequations}
where $\tilde{x}_{t_j,k} = x_{t_{j+k}} - x_{t_j,k}$ is the deviation from the reference trajectory, and $\mu, P, Q, R$ are constant weighting matrices. The first term in~\eqref{equ:obj} penalizes collisions  with obstacles, the second term penalizes the terminal deviation, and the last term penalizes the deviation from the reference trajectory and the control input. Note that in the proposed MPC problem  the penalty on the control input can help reduce the trajectory length. Moreover, the penalty term  $J^{\text{obs}}(x_{t_j, k}, x_{t_j, k+1})$ on the collisions is a point-wise potential function, defined as
\begin{align}
  J^{\text{obs}}(x_{t_j, k}, x_{t_j, k+1}) = \frac{1}{\min_{i} d(o_i, \overline{x_{t_j, k} x_{t_j, k+1}} ) +\epsilon},
\end{align}
where $\min_{i} d(o_i, \overline{x_{t_j, k} x_{t_j, k+1}} )$  is the shortest distance of the line segment that connects waypoints $x_{t_j, k}$ and $x_{t_j, k+1}$, denoted by $\overline{x_{t_j, k} x_{t_j, k+1}}$,  to any obstacle, $\epsilon$ is a small positive constant to avoid the singularity.

\subsection{The Integrated Socially-Aware Planning Algorithm}\label{sec:PlanningAlgorithm}
Alg.~\ref{alg:zero} shows the proposed online solution to Problem~\ref{prob:1} during the whole time period $T_0, T_1, \ldots, T_{K}$, that integrates zeroth-order optimization for social awareness of the desired trajectory and MPC for collision avoidance and dynamic feasibility. As shown in~\cite{zhang2017improved}, if the objective function is non-degenerate, which is a weaker condition than the strong convexity, the dynamic regret in~\eqref{equ:regret} can be improved by estimating the gradient multiple times per time instant. Although, compared to~\cite{zhang2017improved}, here the form of $h_{T_i}$ is unknown, we adopt the same idea and run $N$ gradient updates at every time instant. Note that the inner iteration [line~\ref{zero:inner}] of Alg.~\ref{alg:zero}, reduces to a common online algorithm when $N=1$. Within each inner iteration, the humans are queried for feedback twice: one for the two perturbed trajectories and one for the evaluation of the updated trajectory. Therefore, the number of queries during each time interval $[T_i,T_{i+1})$ is twice  as much as the number of inner iterations. Note that the number of queries can be further reduced by running the $N$ inner iterations without checking whether the updated trajectory has $0$-complaints or not. In this case, humans are only queried once per iteration. In what follows, we discuss Alg.~\ref{alg:zero} in details.

 {A good initial trajectory $\x_{T_0}^0$, starting from the starting location $x_0$ and ending at the  target location $x_g$, can be a collision-free trajectory that is provided by motion planning algorithms, such as RRT$^*$~\cite{karaman2011sampling}. But it  can also be any other trajectory without considering collision since MPC will address this.}
   For each time interval $[T_i, T_{i+1}), \forall i\in \{0, \ldots,  K-1\}$, Alg.~\ref{alg:zero} runs multiple iterations to obtain a trajectory with high social awareness. Specifically, at every iteration $k$, Alg.~\ref{alg:zero} first collects the human feedback on the trajectory $m(\x_{T_i}^{k})$. If $m(\x_{T_i}^k)$ has zero complaints or a predefined maximum number of iterations is reached, then the inner iteration terminates and the trajectory returned by MPC is the solution for the time interval $[T_i, T_{i+1})$ [lines~\ref{zero:tm}-\ref{zero:break}]. Otherwise, Alg.~\ref{alg:zero} updates the waypoints in $\ccalS_p$, perturbs the current iterate $\x_{T_i}^k$ by adding and subtracting $\delta \mathbf{u}^{\ccalS_p}$, and queries the humans for feedback twice [lines~\ref{zero:set2perturb}-\ref{zero:feedback}]. Then, the algorithm computes the gradient for the objective function and updates the iterate [lines~\ref{zero:grad}-\ref{zero:iter}]. We add an extra weight parameter $\alpha>0$ before the gradient of the utility function. At the next time instant $T_{i+1}$, the humans in the environment may move. The trajectory from the previous instant $T_i$ is used as a warm start to initialize $\x_{T_{i+1}}^0$ at instant $T_{i+1}$.


\begin{algorithm}[t]
        \caption{Online trajectory planning based on human feedback}\label{alg:zero}
        \KwIn{Initial trajectory $\x^0_{T_0}$, number of time instants $K$, maximum number of iterations $N$, weights $\alpha, \rho$, exploration parameter $\delta$, step size $\eta$.}
        \For{$0 \leq i \leq K-1$}{
          \For{$0\leq k \leq N$\label{zero:inner}}{
            \uIf{$h_{T_i}(m(\x^{k}_{T_i}))=0 $ or $k = N$ \label{zero:tm}}{
              $\x_{T_i} =  m(\x^{k}_{T_i})$\;
                        \textbf{break} \label{zero:break}\;}
                Get the set $\mathcal{S}_{p}$ of waypoints to perturb  \label{zero:set2perturb}\;
                Get the  human feedback $h_T(m(\x^k_{T_i} + \delta \mathbf{u}^\mathcal{S}_{p}))$ and $h_T(m(\x^k_{T_i} -\delta \mathbf{u}^\mathcal{S}_{p}))$ \label{zero:feedback}\;

                Estimate gradient:\label{zero:grad}
               $$\nabla f_{T_i}(\x_{T_i}^k) \approx \alpha \nabla h_{T_i}^{\mathcal{S}_p}(m(\x_{T_i}^k)) + \rho\nabla e^{\mathcal{S}_p} (\x_{T_i}^k).$$\DontPrintSemicolon\;
                \vspace{-0.4cm}
                Update iterate:   \label{zero:iter}
                $$\x_{T_i}^{k+1} = \x_{T_i}^k - \eta \nabla f_{T_i} (\x^k_{T_i}).$$\DontPrintSemicolon\;
                                \vspace{-0.4cm}
          }
          $\x^0_{T_{i+1}} = \x_{T_i}$\;
        }
\end{algorithm}
\section{Numerical Experiments}\label{sec:simulation}
In this section, we present multiple case studies, implemented using Python 3.6.3 on a computer with a 2.3 GHz Intel Core i5 processor and 8G RAM, that illustrate the efficiency of the proposed algorithm. We consider a square continuous environment $\ccalW$ of size $20 \times 20$ where two square obstacles of size $2 \times 2 $ are located near the center. The starting location of the robot is at $(0, 0)$ and the target location is at $(20, 20)$; see also Fig.~\ref{fig:demo}. The start and target locations are considered fixed throughout the simulations. {We generate a trajectory $\x_0$ that consists of 15 waypoints that connects the initial and target locations; see red arrow path in Fig.~\ref{fig:demo}.} This trajectory constitutes the initial trajectory $\x^0_{T_0}$ for the proposed zeroth-order method.

Furthermore, we consider  a group of $p\in\{20, 30, 40, 50, 60\}$ humans in the environment $\ccalW\setminus \ccalO$. At time $T_0$, we assume that the human population is randomly located in the environment. Then, at each time instant $T_i$, each individual moves by uniformly sampling a location within a circle of radius of $r\in\{0.3, 0.5, 1\}$ centered at their current location. The size of the circle captures the range of an individual's movement. To quantify  human feedback, we assume that each individual possesses a restricted zone and makes a complaint if there exists a trajectory segment passing through the restricted zone. The restricted zone is represented as a circle of radius $r_k$, where $r_k$ is human-dependent and it reflects the sensitivity of an individual to violation of their restricted zone; see also Fig.~\ref{fig:demo}. For every human  $k\in \{1, \ldots,p\}$, we let $r_k$ randomly take a value in the set $\{0.3,  0.4, 0.5, 0.7\}$. The radius of the restricted zone is randomly updated at subsequent time instants $T_i$, which  captures the temporal variation of a human's feedback. We emphasize that the restricted zone is only a construct used to generate complaints and it is not known by the zeroth order planner; neither are the human locations. Finally, we consider a time period containing the sequence of time instants $T_0, \ldots, T_{30}$. The parameters of the zeroth order method in Alg.~\ref{alg:zero} are set as $\alpha = 10, \rho=1, \delta=10$ and $\eta=0.1$ for the full perturbation scheme and $\eta=0.5$ for the local perturbation scheme.
\begin{table}[!t]
  \caption{Statistics for varying number of humans}\label{table:1}
  \begin{tabular}{ccccc}
    \toprule
    $p$   & $N_{\text{full}}$ &  $N_{\text{local}}$ & $d_{\text{full}}$ & $d_{\text{local}}$ \\
    \midrule
    $20$ &  2.35$\pm$2.37 & 3.25$\pm$4.44 & 31.88$\pm$2.42 & 30.34$\pm$0.99 \\
    $30$ &  5.58$\pm$5.19${(1)}$  & 5.20$\pm$8.33 & 34.50$\pm$4.70 & 31.46$\pm$2.84\\
    $40$ & 12.20$\pm$7.67 & 7.75$\pm$4.63 & 39.36$\pm$6.14 & 35.57$\pm$6.51\\
    $50$ & 12.35$\pm$13.10 & 7.60$\pm$8.11 & 39.56$\pm$8.52 & 35.60$\pm$6.18\\
    $60$ & 18.33$\pm$13.17${(5)}$ & 8.68$\pm$8.52${(1)}$ & 43.92$\pm$8.06 & 38.01$\pm$8.23\\
    \bottomrule
  \end{tabular}

  $N_{\text{full}}$ and $N_{\text{local}}$ represent the number of iterations taken by the full and local perturbation schemes, respectively, and $d_{\text{full}}$ and $d_{\text{local}}$ represent the trajectory length returned by full and local perturbation schemes, respectively. The numbers in the parentheses next to certain results report the number of trials out of 20 where no socially-aware trajectory is generated within 50 iterations.  During these trials, paths with high social value were still generated, but the number of complaints was not zero.
\end{table}
\subsection{Robot Model}
We consider a unicycle robot model~\cite{ostafew2016robust}. Let $x = (x^1, x^2) \in \ccalW \subset \mathbb{R}^2$ and $\theta \in (-\pi, \pi]$, respectively, represent the location and orientation of the robot, and $u  = (v, \omega) \in \ccalU$ denote the control input, where $v$ and $\omega$ are the linear and angular velocity, respectively.  Then the discrete kinematic equation governing  the state $q = (x, \theta)$ is
\begin{align}
    q_{t_{j+1}}& =  f (q_{t_j}, u_{t_j})  = q_{t_j} + \Delta t \left [\begin{array}{cc}
        \cos{\theta_{t_j}} & 0 \\
         \sin{\theta_{t_j}} & 0\\
         0 & 1 \\
    \end{array} \right] u_{t_j},
\end{align}
where $\Delta t = t_{j+1} - t_j$. In the simulation experiments that follow,  the MPC parameters are set as $N_m = 5, \mu=50, P=Q=\text{diag}([25, 25]), R=\text{diag}([10,1]), \epsilon=10^{-8}, v\in[0.1, 5], \omega\in(-\pi, \pi], \Delta t =1$.

\subsection{Socially-Aware Planning in Stationary Environments}\label{sec:stationary}
\begin{figure}[!t]
 \centering
  \subfigure[$p = 50$]{
    \label{fig:demo50}
    \includegraphics[width=0.8\linewidth, clip]{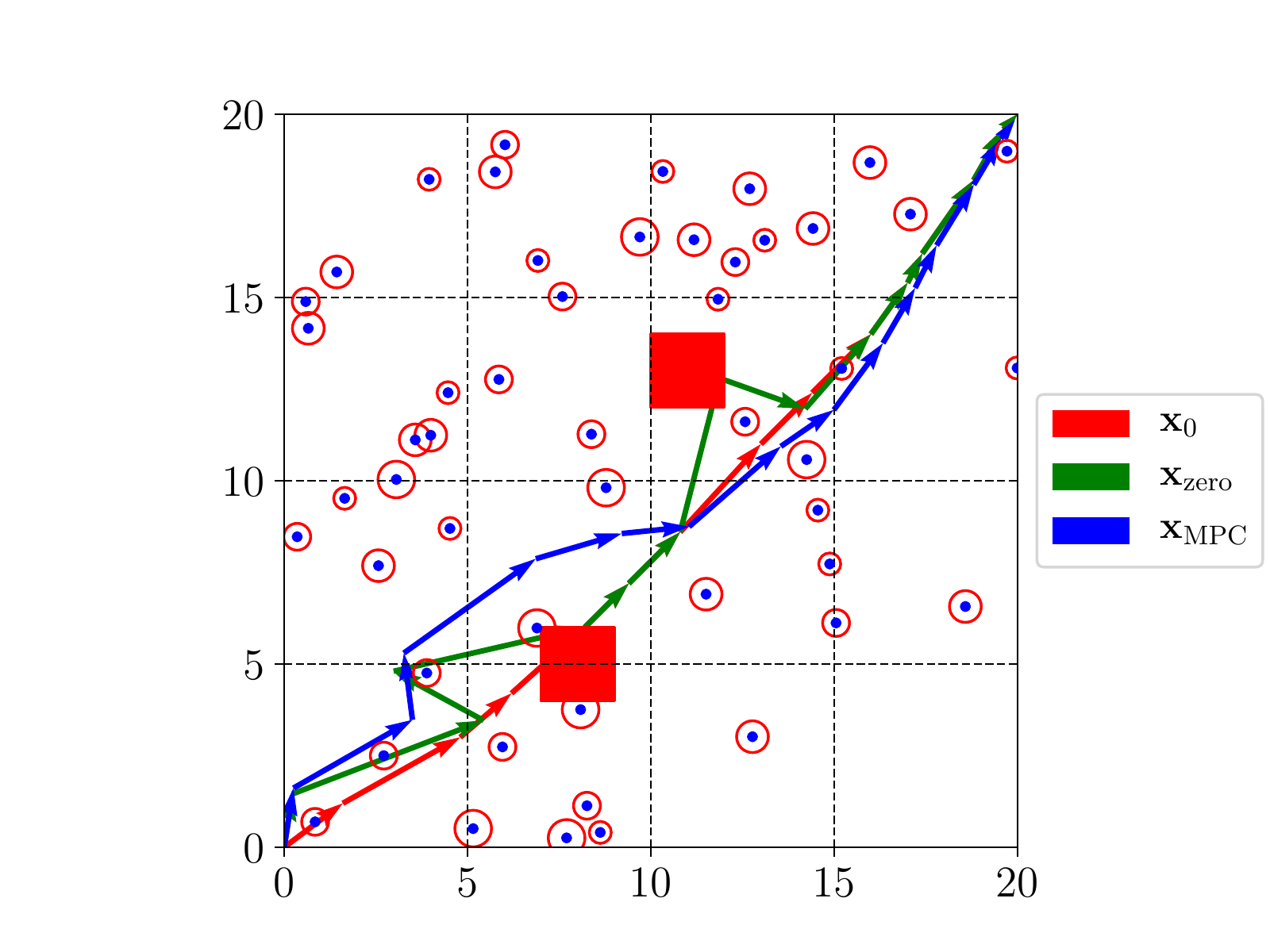}}
  \subfigure[$p = 60$]{
     \label{fig:demo60}
     \includegraphics[width=0.8\linewidth, clip]{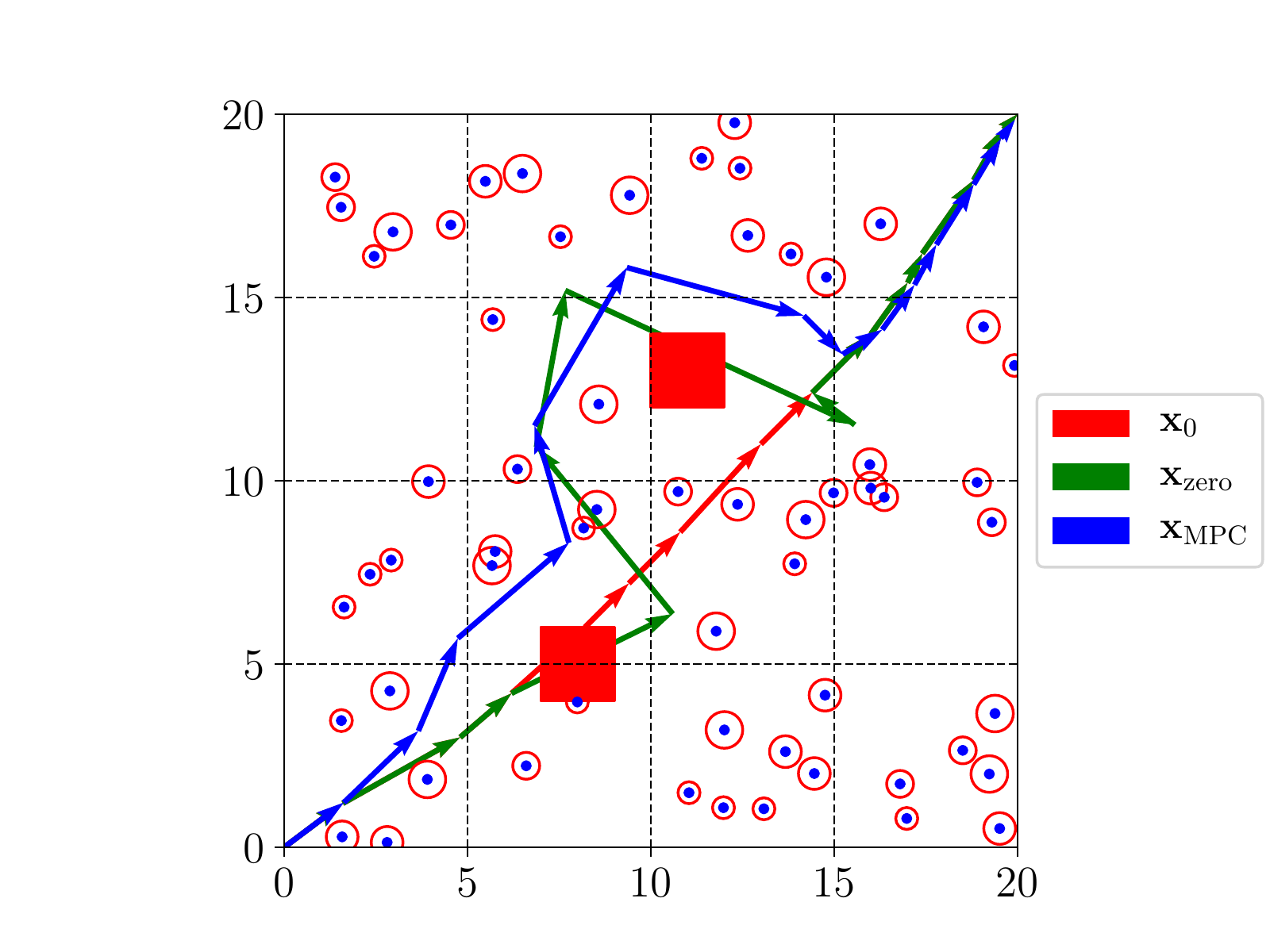}}
  \caption{Sample trajectories. The blue dots represent humans and the red circles are restricted zone. The red squares represent obstacles. The red arrow path is the initial trajectory, and the green and blue arrow path are returned by the zeroth order method and MPC motion planner, respectively.}\label{fig:demo}
\end{figure}

We first validate Alg.~\ref{alg:zero} and compare the full perturbation and local perturbation methods, discussed in Section~\ref{sec:zeroth}, in situations where humans do not move. We set $K=1$ (consider $T_0$ only) and $N= 50$ in Alg.~\ref{alg:zero} and, as discussed in Section~\ref{sec:PlanningAlgorithm},  we perform up to $N$ gradient updates until the inner iteration terminates, with querying humans for feedback on the updated trajectory, so that the total number of queries is twice as much as the number of iterations that Alg.~\ref{alg:zero} takes. Moreover, we use the trajectory $\x_0$ as the initial trajectory $\x^0_{T_0}$ for both full perturbation and local perturbation schemes. The number $p$ of humans varies from 20 to 60. For a specific value of $p$, we report the mean and standard deviation of the number of iterations taken to find a collision-free, dynamically-feasible and socially-aware trajectory, averaged over 20 trials; see also Table~\ref{table:1}. {We also present trajectory length as another metric to compare the full and local perturbation methods.}
Observe that, for each scheme, both the average number of iterations required to obtain a socially-aware trajectory and the trajectory length grow as $p$ increases. In comparison, the local perturbation scheme requires fewer iterations and generates shorter-length trajectories than the full perturbation method. Note that the average number of queries for human feedback, of Alg.~\ref{alg:zero} is low; between $20$ and $40$ times for the local and full perturbation methods, respectively. Therefore, the proposed method can be used in practice without burdening humans with prohibitively many feedback requests. In contrast, the reinforcement learning-based methods discussed in Section~\ref{sec:humanpreference} will require orders of magnitude more samples to obtain an optimal policy. Two sample trajectories returned by the local perturbation scheme, when  $p= 50, 60$, are shown in Fig.~\ref{fig:demo}. We see that the trajectory returned by the MPC motion planner successfully avoids human restricted zone and obstacles, whereas its reference trajectory returned by the zeroth order method violates the restricted zone and collides with obstacles.

\begin{figure*}[!t]
  \centering
  \subfigure[$r=0.3$]{
    \label{fig:full_local_a}
    \includegraphics[width=0.32\linewidth, clip]{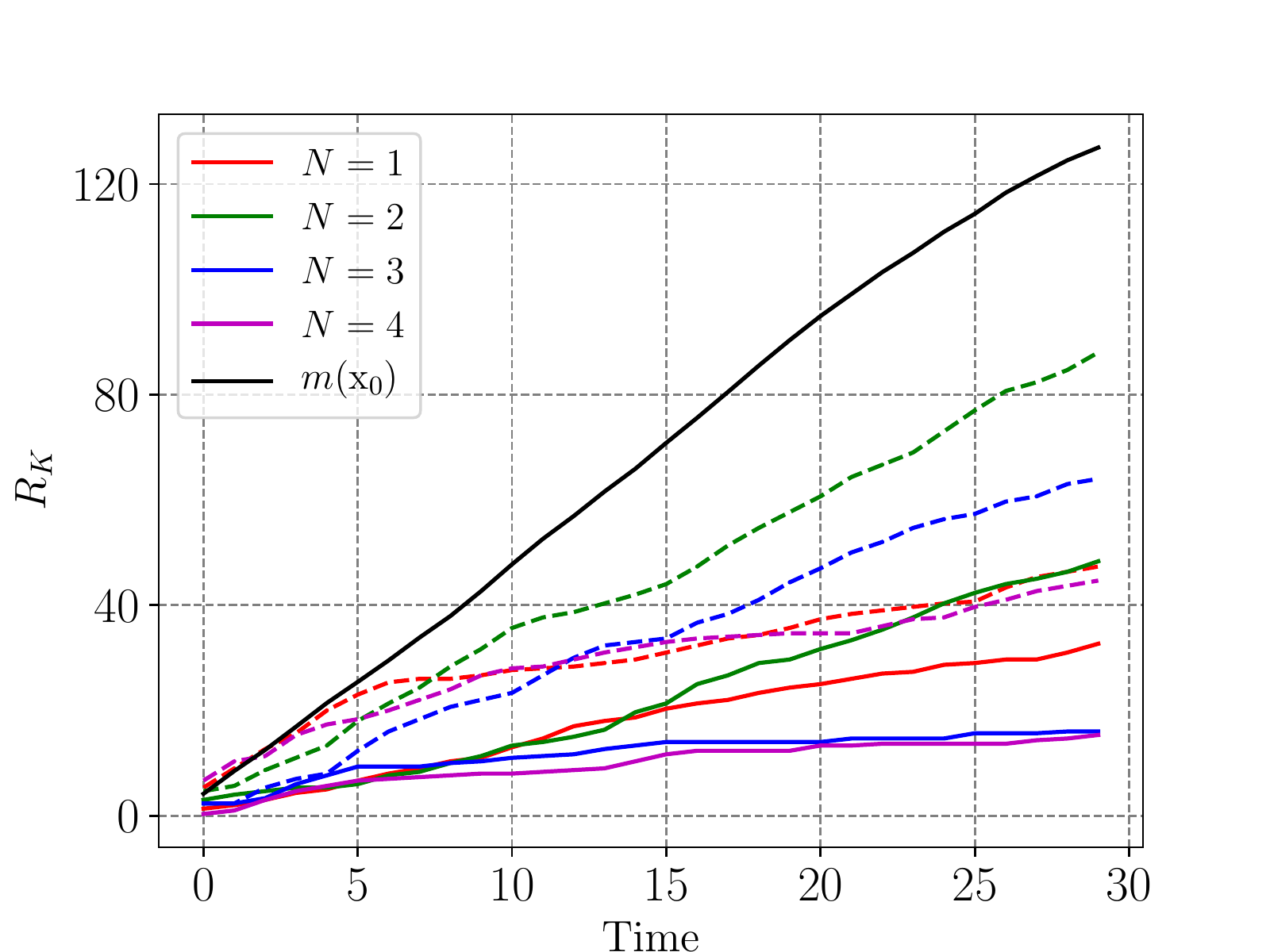}}
  \subfigure[$r=0.5$]{
    \includegraphics[width=0.32\linewidth, clip]{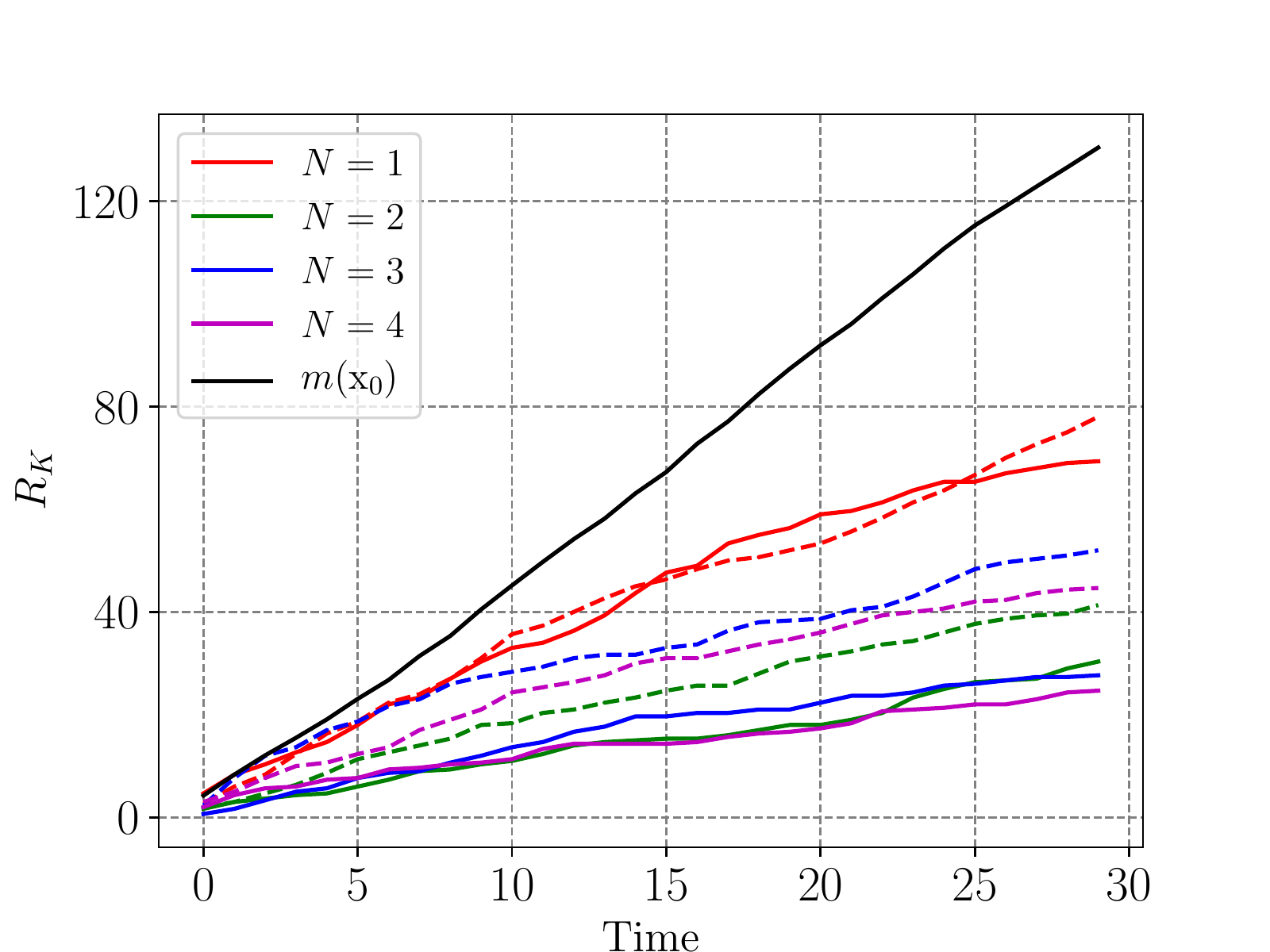}}
   \subfigure[$r=1$]{
    \includegraphics[width=0.32\linewidth, clip]{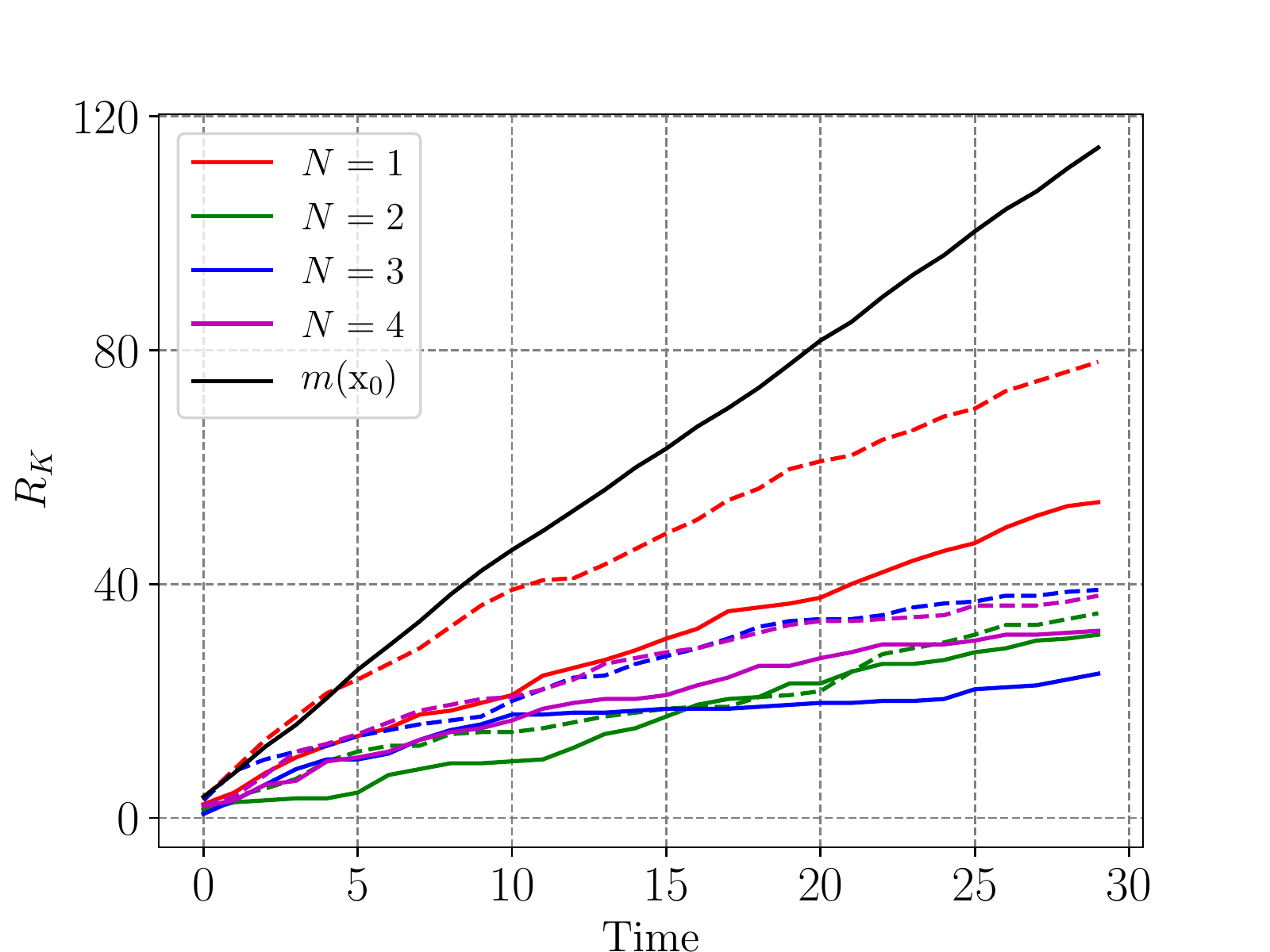}}
  \caption{Cumulative regret with respect to $N$ for different human motion radii $r$. {The $x$-axis corresponds to the index $i$ of time instant $T_i$.} The black solid line is the baseline. The colorred dashed and solid lines represent the cumulative regret achieved by the full and local perturbation schemes, respectively. 
        \label{fig:full_local}}
\end{figure*}
\subsection{Socially-Aware Planning in Dynamic Environments}
In this section, we consider situations where humans move at time instants $T_i, i\in\{0,\ldots, 30\}$ and their corresponding utilities change. Therefore, we solve the online problem~\eqref{equ:minfeedbacpenaltyapprox}. At each time instant, the full and local perturbation schemes are used for the same human distribution and human utility function. 
We assume that there always exists a dynamically-feasible, collision-free and socially-aware trajectory during each time interval. That is, each trajectory in the minimizer sequence ${\x}^*_{T_0}, \ldots, \x^*_{T_{K-1}}$ of problem~\eqref{equ:offline} incurs zero complaints. In this case, the cumulative regret in~\eqref{equ:regret} is reduced to
\begin{align}
 R_K =  \sum_{i=0}^{K-1} h_{T_i}(\tilde{\x}_{T_i})
\end{align}

In the following simulation, we fix the number of humans at 50 and vary the movement radius $r\in\{0.3, 0.5, 1\}$ and the maximum number $N\in\{1,2,3,4\}$ of gradient updates in Alg.~\ref{alg:zero}. Different from~\ref{sec:stationary}, we perform the $N$ gradient updates without querying humans for feedback so that the total number of queries is equal to the number of iterations of Alg.~\ref{alg:zero}. The value or the radius $r$ affects the relative change between human distributions and, therefore, the relative change between  optimization problems at consecutive  time intervals. For each combination of $r$ and $N$, we run 5 trials and report the mean cumulative regret as a function of the time instants $T_i$, for the full and local perturbation schemes, respectively. As a baseline, we also report the mean cumulative regret incurred by the initial trajectory $m(\x_{T_0})$ returned by MPC, at every time interval $[T_i,T_{i+1})$ assuming that this is also the trajectory returned by the algorithm at that time interval. This baseline is  equivalent to the case where $N=0$. Fig.~\ref{fig:full_local} depicts the growth of the regret $R_T$ with respect to the maximum number of iterations $N$ for different human motion radii $r$.  Observe that, compared to the baseline trajectory $m(\x_{T_0})$, the full and local perturbation schemes can improve the social value of the trajectory with few queries. Particularly, using only 3 iterations/queries at each time instant, both methods can provide a good tradeoff between decreasing the cumulative  regret and limiting the human query frequency. Observe that in  Fig.~\ref{fig:full_local_a} for $r=0.3$, the local perturbation scheme performs better than the full perturbation scheme for all values of $N$, but this advantage becomes weak as the human motion radius $r$ increases. This is because as $r$ grows, the relative change in the human distribution becomes large, and so does the the relative change between consecutive optimization problems. In the local perturbation scheme, for each optimization problem, more waypoints of the trajectory returned from the previous time instant have to be modified, due to the larger change in the human distribution  when $r$ increases.


\subsection{Limitations and Future Research Directions}
We have conducted a large number of numerical simulations and have tested the proposed framework in challenging planning problems. One limitation of this framework, in its current form, is that it does not perform well in situations where the target location of the robot is very close to obstacles. This is due to the potential functions used for obstacle avoidance in the MPC motion planner that repel the robot trajectory away from the obstacles. For example, our method has not been able to solve planning problems where the target is located in narrow passages formed by obstacles. Additionally, the proposed framework in its present form can not easily handle complaints, e.g., about robots being too far from humans; in our simulations complaints are about robots being too close to humans. There are a couple of reasons for this. First, the current local perturbation scheme is not applicable since humans are not able to specify a  trajectory segment that affects them if the trajectory is far from them. Second, the full perturbation scheme is also challenging to apply because it is possible that trajectories at consecutive iterations are first close and then far from humans so that it is difficult to estimate zeroth order gradients. The reason for this behavior is the stochastic nature of the full perturbation method. We are currently exploring extensions of our proposed socially-aware planning framework to more complex environments and diverse reasons for human complaints.
\section{Conclusion}
In this paper, we considered the problem of designing
collision-free, dynamically feasible, and socially-aware
trajectories for robots operating in environments populated by
humans. We assumed that humans can provide bandit feedback indicating a complaint or no complaint on the part
of the robot trajectory that caused them discomfort, and
do not reveal any contextual information about their locations or the reasons for their complaints. Assuming that humans can move in the obstacle-free space and, as a result, human utility can change, we formulated this planning problem as an online
optimization problem that minimizes the social value of the
time-varying robot trajectory, defined by the total number of
incurred human complaints. As the human utility is unknown,
we employed zeroth order optimization methods to solve this
problem, which we combined with off-the-shelf motion planners
to satisfy the dynamic feasibility and collision-free properties
of the resulting trajectories. To the best of our knowledge, this
is the first approach to socially-aware robot planning that is
not restricted to avoiding collisions with humans but, instead,
focuses on increasing the social value of the robot trajectories. Compared to data-hungry preference learning methods that can be used to solve similar but stationary socially-aware planning problems, our framework is more practical since it only requires small number of iterations and, therefore, queries for human feedback, to return socially-aware robot paths.
\newcommand{\BIBdecl}{\setlength{\itemsep}{0.25 em}}
\bibliographystyle{IEEEtran}
\bibliography{biblio}

\begin{thebibliography}{10}
\providecommand{\url}[1]{#1}
\csname url@samestyle\endcsname
\providecommand{\newblock}{\relax}
\providecommand{\bibinfo}[2]{#2}
\providecommand{\BIBentrySTDinterwordspacing}{\spaceskip=0pt\relax}
\providecommand{\BIBentryALTinterwordstretchfactor}{4}
\providecommand{\BIBentryALTinterwordspacing}{\spaceskip=\fontdimen2\font plus
\BIBentryALTinterwordstretchfactor\fontdimen3\font minus
  \fontdimen4\font\relax}
\providecommand{\BIBforeignlanguage}[2]{{%
\expandafter\ifx\csname l@#1\endcsname\relax
\typeout{** WARNING: IEEEtran.bst: No hyphenation pattern has been}%
\typeout{** loaded for the language `#1'. Using the pattern for}%
\typeout{** the default language instead.}%
\else
\language=\csname l@#1\endcsname
\fi
#2}}
\providecommand{\BIBdecl}{\relax}
\BIBdecl

\bibitem{kruse2013human}
T.~Kruse, A.~K. Pandey, R.~Alami, and A.~Kirsch, ``Human-aware robot
  navigation: A survey,'' \emph{Robotics and Autonomous Systems}, vol.~61,
  no.~12, pp. 1726--1743, 2013.

\bibitem{chen2017socially}
Y.~F. Chen, M.~Everett, M.~Liu, and J.~P. How, ``Socially aware motion planning
  with deep reinforcement learning,'' in \emph{2017 IEEE/RSJ International
  Conference on Intelligent Robots and Systems (IROS)}.\hskip 1em plus 0.5em
  minus 0.4em\relax IEEE, 2017, pp. 1343--1350.

\bibitem{ferrer2013robot}
G.~Ferrer, A.~Garrell, and A.~Sanfeliu, ``Robot companion: A social-force based
  approach with human awareness-navigation in crowded environments,'' in
  \emph{2013 IEEE/RSJ International Conference on Intelligent Robots and
  Systems}.\hskip 1em plus 0.5em minus 0.4em\relax IEEE, 2013, pp. 1688--1694.

\bibitem{trautman2015robot}
P.~Trautman, J.~Ma, R.~M. Murray, and A.~Krause, ``Robot navigation in dense
  human crowds: Statistical models and experimental studies of human--robot
  cooperation,'' \emph{The International Journal of Robotics Research},
  vol.~34, no.~3, pp. 335--356, 2015.

\bibitem{kuderer2012feature}
M.~Kuderer, H.~Kretzschmar, C.~Sprunk, and W.~Burgard, ``Feature-based
  prediction of trajectories for socially compliant navigation.'' in
  \emph{Robotics: science and systems}, 2012.

\bibitem{kretzschmar2016socially}
H.~Kretzschmar, M.~Spies, C.~Sprunk, and W.~Burgard, ``Socially compliant
  mobile robot navigation via inverse reinforcement learning,'' \emph{The
  International Journal of Robotics Research}, vol.~35, no.~11, pp. 1289--1307,
  2016.

\bibitem{vemula2017modeling}
A.~Vemula, K.~Muelling, and J.~Oh, ``Modeling cooperative navigation in dense
  human crowds,'' in \emph{2017 IEEE International Conference on Robotics and
  Automation (ICRA)}.\hskip 1em plus 0.5em minus 0.4em\relax IEEE, 2017, pp.
  1685--1692.

\bibitem{alahi2016social}
A.~Alahi, K.~Goel, V.~Ramanathan, A.~Robicquet, L.~Fei-Fei, and S.~Savarese,
  ``Social lstm: Human trajectory prediction in crowded spaces,'' in
  \emph{Proceedings of the IEEE conference on computer vision and pattern
  recognition}, 2016, pp. 961--971.

\bibitem{ratliff2009chomp}
N.~Ratliff, M.~Zucker, J.~A. Bagnell, and S.~Srinivasa, ``Chomp: Gradient
  optimization techniques for efficient motion planning,'' 2009.

\bibitem{schulman2013finding}
J.~Schulman, J.~Ho, A.~X. Lee, I.~Awwal, H.~Bradlow, and P.~Abbeel, ``Finding
  locally optimal, collision-free trajectories with sequential convex
  optimization.'' in \emph{Robotics: science and systems}, vol.~9, no.~1.\hskip
  1em plus 0.5em minus 0.4em\relax Citeseer, 2013, pp. 1--10.

\bibitem{toussaint2017tutorial}
M.~Toussaint, ``A tutorial on newton methods for constrained trajectory
  optimization and relations to slam, gaussian process smoothing, optimal
  control, and probabilistic inference,'' in \emph{Geometric and numerical
  foundations of movements}.\hskip 1em plus 0.5em minus 0.4em\relax Springer,
  2017, pp. 361--392.

\bibitem{ratliff2015understanding}
N.~Ratliff, M.~Toussaint, and S.~Schaal, ``Understanding the geometry of
  workspace obstacles in motion optimization,'' in \emph{2015 IEEE
  International Conference on Robotics and Automation (ICRA)}.\hskip 1em plus
  0.5em minus 0.4em\relax IEEE, 2015, pp. 4202--4209.

\bibitem{webb2013kinodynamic}
D.~J. Webb and J.~Van Den~Berg, ``Kinodynamic rrt*: Asymptotically optimal
  motion planning for robots with linear dynamics,'' in \emph{2013 IEEE
  International Conference on Robotics and Automation}.\hskip 1em plus 0.5em
  minus 0.4em\relax IEEE, 2013, pp. 5054--5061.

\bibitem{kim2015trajectory}
J.-J. Kim and J.-J. Lee, ``Trajectory optimization with particle swarm
  optimization for manipulator motion planning,'' \emph{IEEE Transactions on
  Industrial Informatics}, vol.~11, no.~3, pp. 620--631, 2015.

\bibitem{kalakrishnan2011stomp}
M.~Kalakrishnan, S.~Chitta, E.~Theodorou, P.~Pastor, and S.~Schaal, ``Stomp:
  Stochastic trajectory optimization for motion planning,'' in \emph{2011 IEEE
  international conference on robotics and automation}.\hskip 1em plus 0.5em
  minus 0.4em\relax IEEE, 2011, pp. 4569--4574.

\bibitem{mcgill2016low}
S.~G. McGill, S.-J. Yi, and D.~D. Lee, ``Low dimensional human preference
  tracking for motion optimization,'' in \emph{2016 IEEE International
  Conference on Robotics and Automation (ICRA)}.\hskip 1em plus 0.5em minus
  0.4em\relax IEEE, 2016, pp. 2867--2872.

\bibitem{jain2015learning}
A.~Jain, S.~Sharma, T.~Joachims, and A.~Saxena, ``Learning preferences for
  manipulation tasks from online coactive feedback,'' \emph{The International
  Journal of Robotics Research}, vol.~34, no.~10, pp. 1296--1313, 2015.

\bibitem{menon2015towards}
A.~Menon, P.~Kacker, and S.~Chitta, ``Towards a data-driven approach to human
  preferences in motion planning,'' in \emph{2015 IEEE International Conference
  on Robotics and Automation (ICRA)}.\hskip 1em plus 0.5em minus 0.4em\relax
  IEEE, 2015, pp. 920--927.

\bibitem{daniel2015active}
C.~Daniel, O.~Kroemer, M.~Viering, J.~Metz, and J.~Peters, ``Active reward
  learning with a novel acquisition function,'' \emph{Autonomous Robots},
  vol.~39, no.~3, pp. 389--405, 2015.

\bibitem{wirth2017survey}
C.~Wirth, R.~Akrour, G.~Neumann, and J.~F{\"u}rnkranz, ``A survey of
  preference-based reinforcement learning methods,'' \emph{The Journal of
  Machine Learning Research}, vol.~18, no.~1, pp. 4945--4990, 2017.

\bibitem{wilson2012bayesian}
A.~Wilson, A.~Fern, and P.~Tadepalli, ``A bayesian approach for policy learning
  from trajectory preference queries,'' in \emph{Advances in neural information
  processing systems}, 2012, pp. 1133--1141.

\bibitem{busa2014preference}
R.~Busa-Fekete, B.~Sz{\"o}r{\'e}nyi, P.~Weng, W.~Cheng, and E.~H{\"u}llermeier,
  ``Preference-based reinforcement learning: evolutionary direct policy search
  using a preference-based racing algorithm,'' \emph{Machine Learning},
  vol.~97, no.~3, pp. 327--351, 2014.

\bibitem{furnkranz2012preference}
J.~F{\"u}rnkranz, E.~H{\"u}llermeier, W.~Cheng, and S.-H. Park,
  ``Preference-based reinforcement learning: a formal framework and a policy
  iteration algorithm,'' \emph{Machine learning}, vol.~89, no. 1-2, pp.
  123--156, 2012.

\bibitem{christiano2017deep}
P.~F. Christiano, J.~Leike, T.~Brown, M.~Martic, S.~Legg, and D.~Amodei, ``Deep
  reinforcement learning from human preferences,'' in \emph{Advances in Neural
  Information Processing Systems}, 2017, pp. 4299--4307.

\bibitem{sadigh2017active}
D.~Sadigh, A.~D. Dragan, S.~Sastry, and S.~A. Seshia, ``Active preference-based
  learning of reward functions.'' in \emph{Robotics: Science and Systems},
  2017.

\bibitem{pinsler2018sample}
R.~Pinsler, R.~Akrour, T.~Osa, J.~Peters, and G.~Neumann, ``Sample and feedback
  efficient hierarchical reinforcement learning from human preferences,'' in
  \emph{2018 IEEE International Conference on Robotics and Automation
  (ICRA)}.\hskip 1em plus 0.5em minus 0.4em\relax IEEE, 2018, pp. 596--601.

\bibitem{nesterov2017random}
Y.~Nesterov and V.~Spokoiny, ``Random gradient-free minimization of convex
  functions,'' \emph{Foundations of Computational Mathematics}, vol.~17, no.~2,
  pp. 527--566, 2017.

\bibitem{gasnikov2017stochastic}
A.~V. Gasnikov, E.~A. Krymova, A.~A. Lagunovskaya, I.~N. Usmanova, and F.~A.
  Fedorenko, ``Stochastic online optimization. single-point and multi-point
  non-linear multi-armed bandits. convex and strongly-convex case,''
  \emph{Automation and remote control}, vol.~78, no.~2, pp. 224--234, 2017.

\bibitem{shahrampour2017distributed}
S.~Shahrampour and A.~Jadbabaie, ``Distributed online optimization in dynamic
  environments using mirror descent,'' \emph{IEEE Transactions on Automatic
  Control}, vol.~63, no.~3, pp. 714--725, 2017.

\bibitem{mokhtari2016online}
A.~Mokhtari, S.~Shahrampour, A.~Jadbabaie, and A.~Ribeiro, ``Online
  optimization in dynamic environments: Improved regret rates for strongly
  convex problems,'' in \emph{2016 IEEE 55th Conference on Decision and Control
  (CDC)}.\hskip 1em plus 0.5em minus 0.4em\relax IEEE, 2016, pp. 7195--7201.

\bibitem{rawlings2000tutorial}
J.~B. Rawlings, ``Tutorial overview of model predictive control,'' \emph{IEEE
  control systems magazine}, vol.~20, no.~3, pp. 38--52, 2000.

\bibitem{grune2017nonlinear}
L.~Gr{\"u}ne and J.~Pannek, ``Nonlinear model predictive control,'' in
  \emph{Nonlinear Model Predictive Control}.\hskip 1em plus 0.5em minus
  0.4em\relax Springer, 2017, pp. 45--69.

\bibitem{allgower2012nonlinear}
F.~Allg{\"o}wer and A.~Zheng, \emph{Nonlinear model predictive control}.\hskip
  1em plus 0.5em minus 0.4em\relax Birkh{\"a}user, 2012, vol.~26.

\bibitem{lapierre2007combined}
L.~Lapierre, R.~Zapata, and P.~Lepinay, ``Combined path-following and obstacle
  avoidance control of a wheeled robot,'' \emph{The International Journal of
  Robotics Research}, vol.~26, no.~4, pp. 361--375, 2007.

\bibitem{yoon2009model}
Y.~Yoon, J.~Shin, H.~J. Kim, Y.~Park, and S.~Sastry, ``Model-predictive active
  steering and obstacle avoidance for autonomous ground vehicles,''
  \emph{Control Engineering Practice}, vol.~17, no.~7, pp. 741--750, 2009.

\bibitem{gao2011predictive}
Y.~Gao, T.~Lin, F.~Borrelli, E.~Tseng, and D.~Hrovat, ``Predictive control of
  autonomous ground vehicles with obstacle avoidance on slippery roads,'' in
  \emph{ASME 2010 dynamic systems and control conference}.\hskip 1em plus 0.5em
  minus 0.4em\relax American Society of Mechanical Engineers Digital
  Collection, 2011, pp. 265--272.

\bibitem{zhang2017improved}
L.~Zhang, T.~Yang, J.~Yi, J.~Rong, and Z.-H. Zhou, ``Improved dynamic regret
  for non-degenerate functions,'' in \emph{Advances in Neural Information
  Processing Systems}, 2017, pp. 732--741.

\bibitem{karaman2011sampling}
S.~Karaman and E.~Frazzoli, ``Sampling-based algorithms for optimal motion
  planning,'' \emph{The international journal of robotics research}, vol.~30,
  no.~7, pp. 846--894, 2011.

\bibitem{ostafew2016robust}
C.~J. Ostafew, A.~P. Schoellig, and T.~D. Barfoot, ``Robust constrained
  learning-based nmpc enabling reliable mobile robot path tracking,'' \emph{The
  International Journal of Robotics Research}, vol.~35, no.~13, pp. 1547--1563,
  2016.

\end{thebibliography}

\end{document}